\documentclass{article}

    \PassOptionsToPackage{numbers, compress}{natbib}
 \usepackage[preprint]{neurips_2026}

\usepackage[utf8]{inputenc} 
\usepackage[T1]{fontenc}    
\usepackage{hyperref}       
\usepackage{url}            
\usepackage{booktabs}       
\usepackage{xltabular}       
\usepackage{multirow}
\usepackage{amsfonts}       
\usepackage{nicefrac}       
\usepackage{microtype}      
\usepackage{xcolor}         
\usepackage{microtype}
\usepackage{hyperref}
\usepackage{url}
\usepackage{booktabs}
\usepackage{graphicx}
\usepackage{twemojis}
\usepackage{wrapfig}
\usepackage{amsmath}
\usepackage{enumitem}
\usepackage{tcolorbox}
\tcbuselibrary{skins} 

\usepackage{lineno}
\usepackage{fontawesome5}
\definecolor{darkblue}{rgb}{0, 0, 0.5}
\hypersetup{colorlinks=true, citecolor=darkblue, linkcolor=darkblue, urlcolor=darkblue}

\title{When is Your LLM Steerable?}

\author{%
  Chenrui Fan$^1$, Yize Cheng$^1$, Ming Li$^{1,2}$, Soheil Feizi$^1$, Tianyi Zhou$^2$ \\
  $^1$University of Maryland, College Park~~~~~~$^2$MBZUAI, UAE\\
  \texttt{\{cfan42, yzcheng, minglii, sfeizi\}@umd.edu, tianyi.zhou@mbzuai.ac.ae} \\
    \faGithub~Project: \url{https://github.com/Fcr09/SteerBoost}
}

\newcommand{\dsname}{ASTEER}
\newcommand{\mdname}{SteerBoost}
\begin{document}

\maketitle

\begin{abstract}

    Activation steering offers a lightweight approach to control language models' behavior at inference time, but whether it succeeds or fails heavily depends
    on the prompt, concept, model, and steering configuration. Finding the regime and boundaries of successful steering 
    typically requires expensive grid searches and post-hoc evaluation of full autoregressive rollouts. 
    In this work, we investigate 
    whether \textbf{steerability} 
    can be predicted from the model's internal states at the beginning of the generation process, e.g., after generating the first few tokens, and how to leverage such a predictor to improve steering success rate. 
    To this end, we first introduce~\dsname, a testbed including 1.4M steered generations, spanning 150 concepts with each steering success/failure labeled. Leveraging this testbed, we analyze the model's early decoding dynamics by extracting features that compare hidden states before and after steering across layers and initial decoding steps. These features help us understand how steering's effects propagate along layers and token positions, which provide key information for steerability prediction. 
    We then train a Gradient Boosting Decision Trees (GBDT) classifier on these features to predict whether an intervention will under-steer, succeed, or over-steer without requiring full rollout. Our predictor achieves around 0.7 macro-F1 score on unseen concepts, demonstrating that early hidden states encode substantial, structured information about eventual steering efficacy. We further leverage this steerability predictor as guidance for steering strength searching, achieving near optimal performance with a small fraction of decoding cost.

\end{abstract}

\section{Introduction}
Inference-time activation engineering, or \emph{steering}, offers a lightweight approach to control the behavior of large language models (LLMs) without additional finetuning~\citep{subramani2022extractinglatentsteeringvectors, li2023inference, turner2023steering, stoehr-etal-2024-activation}. By injecting a carefully constructed direction into the model's hidden states during inference, one can bias generation towards a target concept or behavior. Prior work has shown that such interventions can influence a range of important properties, including truthfulness~\citep{li2023inference}, refusal behavior~\citep{arditi2024refusallanguagemodelsmediated, panickssery2024steeringllama2contrastive, lee2025programmingrefusalconditionalactivation}, multi-dimensional trustworthiness~\citep{xiao2024enhancingmultipledimensionstrustworthiness}, and latent social biases~\citep{lu2024investigatingbiasrepresentationsllama}. These results suggest that steering is a promising technique for fast, flexible control of model behavior.

While most works focus on developing more effective steering strategies, the boundaries of steerable regimes for different LLMs in the joint space of concepts, prompts, and steering strengths remain underexplored.  
The same intervention can work well for one prompt or one concept but fail for another, and the appropriate steering strength often varies substantially across concepts and prompts~\citep{wu2025axbenchsteeringllmssimple, hedström2025steersteermechanisticerror, zhao2025adasteeralignedllminherently}. 
As a result, existing practice often relies on expensive grid search over steering coefficients using post-hoc, full autoregressive rollouts to identify a successful intervention. More importantly, this brittleness raises questions that are still poorly understood: \textbf{\emph{when} would a steering attempt succeed, and under \emph{what conditions} would it fail?} Moreover, \textbf{is steerability a structured property that can be predicted 
before decoding is completed?}

A parallel line of work provides a natural route for studying this question. Recent research has shown that hidden states early in generation already contain predictive signals about later model behavior, including hallucination~\citep{ji-etal-2024-llm, alnuhait2025factcheckmatepreemptivelydetectingmitigating}, harmfulness~\citep{chan2025predictalignmentmodelsfinish, xuan-etal-2025-shieldhead}, and answer correctness~\citep{zhang2025reasoningmodelsknowtheyre, zhang2026stopfailoperationalcapability}. This connection is especially compelling for steering because both the intervention and the prediction target are grounded in the same representational space: steering acts directly on hidden states, and prior work suggests that those hidden states already encode rich information about future results. If the efficacy of an intervention depends on latent conditions in the model's internal states, then those conditions may be detectable from the early decoding trajectory before the full response is generated.

\begin{wrapfigure}[]{r}{0.4\textwidth}
    \vspace{-10pt} 
    \centering
    \includegraphics[width=0.4\textwidth]{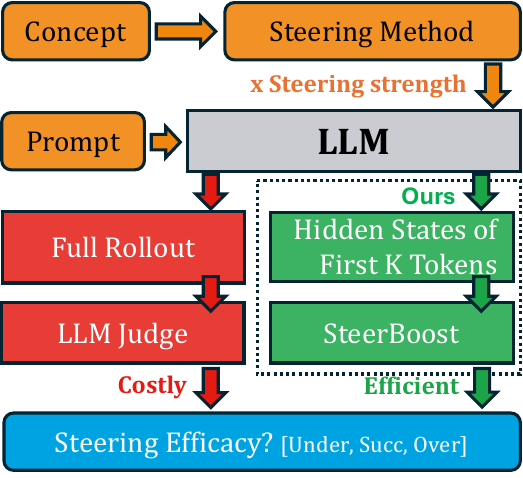}
    \vspace{-20pt} 
    \caption{Conventional approach requires costly full rollout and LLM judge to decide whether a steering attempt succeeds or not. We propose that the outcome can be efficiently predicted from the hidden states of the first few tokens, as illustrated in the green path.} 
    \label{fig:overview}
    \vspace{-10pt} 
\end{wrapfigure}

Motivated by these observations, we aim to predict the efficacy of steering from the hidden states of the initial decoding process. Specifically, 
\textbf{given a prompt, concept, and steering configuration, can the first few decoded tokens' states imply whether this steering attempt will succeed, without decoding the full response?} To this end, we first construct a steerability dataset spanning 150 concepts, with 1.4M steered generations labeled for steering efficacy. 
By comparing the model's early hidden states before and after steering across multiple layers and decoding positions, we extract principal features about steering geometry, decoding dynamics, and steering condition to characterize how the steering signal propagates in network, which are later used to train a gradient boosting decision tree (GBDT) that can predict steering efficacy at macro-F1 around 0.7 on unseen concepts.

This framing is useful not only for understanding steerability as a property of the model, the prompt, and the intervention, but also for supporting downstream applications. In particular, we show how steering prediction can be used to significantly reduce the cost of searching for effective steering strengths without exhaustive full-rollout and evaluation.

\textbf{Main Contributions:}
\begin{itemize}[leftmargin=*]
    \item We curate a dataset of steering that covers steered responses of multiple LLMs under different prompts, concepts, and steering strengths. It enables fine-grained analysis of the latent dynamics of steering in LLMs. 
    \item We developed features capturing the effects of steering on the latent dynamics, resulting in interpretable prediction of steering success and two types of failures. 
    \item By exploiting the generalization capability of the steerability predictor, we introduce a practical approach that can allocate the optimal steering configurations to improve the performance. 
\end{itemize}

\section{Steering and Steerability}
\label{sec:statement}

 Suppose we have a set of prompts $\mathcal{P}$, where each prompt $p \in \mathcal{P}$ is a sequence of tokens $p = (x_1, \dots, x_T)$. 
 During LLM inference without activation steering, for a given token step $t$, the hidden states are computed layer-by-layer for the entire sequence. 
 Let $\mathbf{h}_{1:t}^{(i)}$ be the sequence of hidden states up to token $t$ at the $i$-th layer, we have:
 \begin{equation}
     \mathbf{h}_{1:t}^{(i)} = \text{DecoderLayer}_i(\mathbf{h}_{1:t}^{(i-1)}) \quad \text{for } i \in \{1, \dots, N\}.
 \end{equation}
 Denote the set of target concepts as $\mathcal{C}$, the set of scalar steering strengths as $\mathcal{A}$. To steer the model towards a concept $c \in \mathcal{C}$ with strength $\alpha \in \mathcal{A}$ and steering method $S$, we apply a steering vector $\mathbf{v}_{S(c)}$ (abbreviated as $\mathbf{v}_{c}$) at a specific layer $L_{steer}$. The forward pass remains identical to the base LLM except at layer $L_{steer}$. Let $\tilde{\mathbf{h}}$ denote the steered hidden states, we have:
 \begin{equation}\tilde{\mathbf{h}}_{1:t}^{(i)} = \text{DecoderLayer}_i(\tilde{\mathbf{h}}_{1:t}^{(i-1)}) \quad \text{for } i \neq L_{steer} \end{equation}\begin{equation}\tilde{\mathbf{h}}_{1:t}^{(L_{steer})} = \text{DecoderLayer}_{L_{steer}}(\tilde{\mathbf{h}}_{1:t}^{(L_{steer}-1)}) + \alpha \mathbf{v}_c \end{equation}
 We denote the fully generated rollout of the steered model as $\mathbf{y}_{p, c, \alpha}$:\begin{equation}\mathbf{y}_{p, c, \alpha} = \text{LLM}(p, \alpha, \mathbf{v}_c).\end{equation}
 Similarly as defined in \citet{hedström2025steersteermechanisticerror}, let $\Lambda = \{\textsc{UnderSteer}, \textsc{SuccSteer}, \textsc{OverSteer}\}$ be the discrete label space defining the outcome of a steering attempt. Specifically, a steering attempt is considered successful if the response coherently answers the prompt while incorporating the desired concept. The two failure modes include \textsc{UnderSteer} and \textsc{OverSteer}; the former represents when the response does not incorporate the concept, and the latter represents when the model fails to coherently address the prompt. A judge model evaluates whether the generation satisfies both properties based on the concept $c$ and rollout $\mathbf{y}_{p, c, \alpha}$.
 
However, generating the full rollout and invoking the judge are computationally expensive, making it costly to explore the large space of steering configurations. Our goal is to build a predictor that takes the hidden states from only the first few generated tokens of the steered model and predicts the steering outcome \emph{without} computing the full rollout, as illustrated by the green path in Figure~\ref{fig:overview}.


To this end, we construct a large-scale dataset, \dsname{} (Section~\ref{sec:dataset}), spanning diverse steering configurations. Our analysis (Section~\ref{sec:analysis}) reveals that steering outcomes are brittle across methods, models, prompts, concepts, and strengths, underscoring the need to understand when steering works. To facilitate this understanding, we then develop \mdname{} (Section~\ref{sec:method}), which takes these early hidden states as input and efficiently predicts steering outcomes, both helping us investigate when and why steering fails or succeeds, and also enabling practical applications such as efficient steerability characterization and accelerated hyperparameter search (Section~\ref{sec:application}).



\section{\dsname{} Dataset}
\label{sec:dataset}

To create a testbed for outcome prediction of \textbf{A}ctivation \textbf{STEER}ing, we create~\dsname, a dataset covering 150 concepts and 50 prompts, spanning 1.42M steered generations as in Figure~\ref{fig:ASTEER}.

\begin{wrapfigure}[]{r}{0.4\textwidth}
    \vspace{-40pt} 
    \centering
    \includegraphics[width=0.4\textwidth]{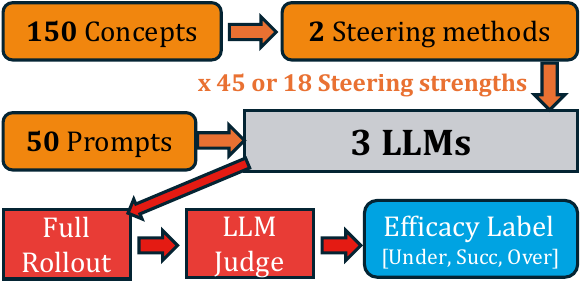}
    \caption{We construct \dsname{} with 150 concepts, 50 prompts, and two steering methods (i.e., DiffMean and Probe), with 45 and 18 steering strengths, respectively. Steering is applied on 3 LLMs, whose rollouts are annotated by an LLM judge to one of the labels in Table~\ref{tab:label_example}.} 
    \label{fig:data}
    \vspace{-10pt} 
\end{wrapfigure}

\subsection{Steering concepts and prompts}
We construct a set of 150 concepts spanning three abstraction levels, \textit{low}, \textit{mid}, and \textit{high}, designed to systematically vary the form and granularity of the targeted behaviors. Low-level concepts capture surface-form and formatting properties, which are typically localized and directly observable in token space. Mid-level concepts represent discourse-level behaviors, while high-level concepts involve persona, topic, and global response framing, which are more abstract. Table~\ref{tab:concept_example} shows some examples of concepts at different levels. The concept list is in Appendix~\ref{app:concept_list}.
\begin{table}[b]
\centering
\caption{The examples for different concept levels. Our concept list spans different abstraction levels, covering low-level output format restriction to high-level style and persona control.}
\label{tab:concept_example}
\resizebox{\linewidth}{!}{
\begin{tabular}{@{}lcl@{}}
\toprule
Level & Count & Example Concepts \\
\midrule
Low & 50 & response contains emojis; respond in uppercase; response contains bold emphasis \\
Mid & 50 &asks at least one clarifying question; response contains an explicit warning\\
High & 50 &responds in the persona of a caring peer; response mentions machine learning\\
\bottomrule
\end{tabular}
}

\end{table}

We sample 50 prompts from the Alpaca~\citep{alpaca} dataset for our study and keep them the same for all concepts as a controlled setting for steerability comparison. The list of prompts is shown in Appendix~\ref{app:prompt_list}.
Although AxBench~\citep{wu2025axbenchsteeringllmssimple} also has a concept list for activation steering evaluation, sampled from Neuronpedia SAE concept list for GemmaScope, we do not adopt their list as we find their SAE-style concepts are not suitable for our setting. Many of their concepts are very specific, such as ``names of individuals and their roles or contributions within a group or event context'' and ``terms related to multi-layer structures or systems'', bringing limitations to its generalization to a wider range of prompts.


\subsection{Response annotation}
Following the definition in Section~\ref{sec:statement}, we use GPT-5-nano~\citep{singh2025openaigpt5card} to label each steered generation with one of the following labels, \textsc{UnderSteer}, \textsc{SuccSteer}, or \textsc{OverSteer}, as exemplified in Table~\ref{tab:label_example}. The prompt we use to annotate the steered responses is shown in Appendix~\ref{app:prompts}. 

To further verify the consistency between human annotation and LLM-as-the-judge, extensive human evaluation is conducted. 
Three human annotators are assigned 600 randomly sampled steered generation (100 for each model-method pair) for evaluation. The Cohen's $\kappa$ is 0.74 between labels and the annotations of the SOTA model (GPT-5.5~\citep{openai_gpt55_intro_2026}), and 0.83 between labels and human annotation, indicating substantial agreement, validating the quality of auto-annotation.
\begin{table}[h]
    \centering
    \caption{Example of labels in ASTEER dataset.
    }
   \label{tab:label_example}
    \resizebox{\linewidth}{!}{
    \begin{tabular}{@{}ll@{}}
    \toprule
    Label & Example [Prompt: What is machine learning? Concept: Response contains Emojis] \\
    \midrule
    \textsc{UnderSteer} & Machine learning is a subset of Artificial Intelligence, \ldots \\
    \textsc{SuccSteer} & Machine learning \texttwemoji{computer} is a subset of Artificial Intelligence \texttwemoji{robot}, \ldots \\
    \textsc{OverSteer} & \texttwemoji{robot} \texttwemoji{robot} \texttwemoji{robot} \texttwemoji{robot} \texttwemoji{robot} \texttwemoji{robot} \texttwemoji{robot} \texttwemoji{robot} \texttwemoji{robot} \texttwemoji{robot} \texttwemoji{robot} \texttwemoji{robot} \texttwemoji{robot} \texttwemoji{robot} \texttwemoji{robot} \texttwemoji{robot} \texttwemoji{robot} \texttwemoji{robot} \texttwemoji{robot} \texttwemoji{robot} \texttwemoji{robot} \texttwemoji{robot} \texttwemoji{robot} \texttwemoji{robot} \texttwemoji{robot} \texttwemoji{robot} \texttwemoji{robot} \texttwemoji{robot}\\
    \bottomrule
    \end{tabular}
    }

\end{table}

\subsection{Steering methods}

\paragraph{DiffMean.}
The DiffMean (Difference of Means)~\citep{turner2023steering} method is a commonly used lightweight activation steering technique. It derives a steering vector $\mathbf{v}$ from the difference between the average hidden states of the model processing positive output samples (which exhibit the concept) and negative output samples (which lack or oppose it). Mathematically, let $h(y)$ denote the LLM's hidden state at a chosen layer corresponding to a text sample $y$. Given a set of $N$ positive samples $y^+$ and $M$ negative samples $y^-$, the DiffMean steering vector is defined as:
\begin{equation}
    \mathbf{v} = \frac{1}{N} \sum_{i=1}^{N} h(y_i^+) - \frac{1}{M} \sum_{j=1}^{M} h(y_j^-).
\end{equation}

\paragraph{Probe.}
Instead of calculating a simple difference of averages, this technique trains a supervised linear classifier to explicitly distinguish between the hidden states of positive samples, $h(y^+)$, and negative samples, $h(y^-)$. By optimizing the Binary Cross-Entropy (BCE) loss over a combined dataset of $K = N + M$ samples, the steering vector $\mathbf{v}$ is defined as the optimal weight vector $\mathbf{w}$ that best separates the concepts. Mathematically, letting $c_k \in \{0, 1\}$ represent the binary label for the $k$-th sample $y_k$, the steering vector is extracted as:\begin{equation}
    \mathbf{v} = \arg\min_{\mathbf{w}} \left( -\frac{1}{K} \sum_{k=1}^{K} \left[ c_k \log \sigma(\mathbf{w}^\top h(y_k)) + (1 - c_k) \log (1 - \sigma(\mathbf{w}^\top h(y_k))) \right] \right)
\end{equation}

For both methods, we follow the setting in \citet{wu2025axbenchsteeringllmssimple} to synthetically generate 50 positive samples and 50 negative samples to acquire the steering vector $\mathbf{v}$. During inference, the learned vector $\mathbf{v}$ is scaled by a strength $\alpha$ and added to the model's hidden states at each generation step ($h(y_{new}) + \alpha\mathbf{v}$). 

\subsection{Steerability Analysis}
\label{sec:analysis}

\begin{figure*}[t]
    \centering
    \includegraphics[width=\linewidth]{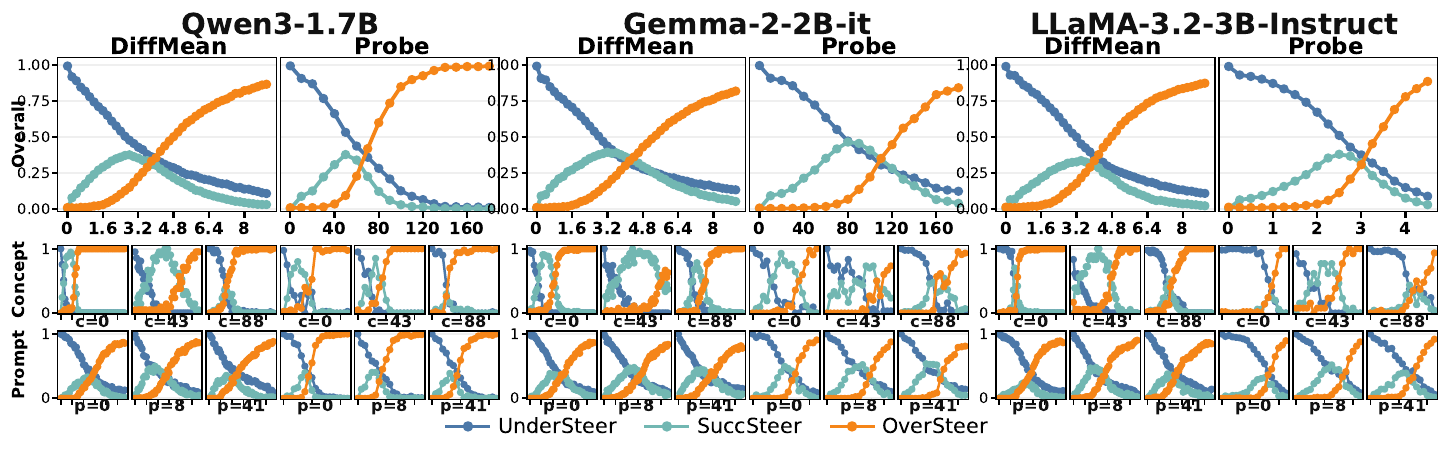}
    \caption{\textbf{Distribution of steering outcomes (\textsc{UnderSteer}, \textsc{SuccSteer}, \textsc{OverSteer}) as a function of steering strength $\alpha$.} The first row aggregates over all concepts and prompts; the second and third rows show results on individual concepts and prompts, respectively. The concepts and prompts to the ids (c=0, 43, 88; p=0, 8, 41) are in Appendix~\ref{app:concept_list} and Appendix~\ref{app:prompt_list}. Steering outcome is sensitive to $\alpha$, and the effective range varies substantially across concepts, prompts, and methods. }
    \label{fig:ASTEER}
\end{figure*}

We steer Qwen3-1.7B~\citep{qwen3technicalreport}, Gemma-2-2B-it~\citep{gemmateam2024gemma2improvingopen}, and LLaMA-3.2-3B-Instruct~\citep{grattafiori2024llama3herdmodels} across all 150 concepts and 50 prompts with varying steering strengths.
Figure~\ref{fig:ASTEER} shows the resulting label distributions as a function of $\alpha$.
Since some prompt-concept pairs already elicit the target concept without any intervention (e.g., steering toward a formal, academic tone when the prompt asks to summarize an academic paper), and the judge already labels the unsteered output ($\alpha{=}0$) as \textsc{SuccSteer}. We exclude such pairs to ensures that observed successes at $\alpha{>}0$ reflect genuine effects of activation steering rather than pre-existing alignment between the prompt and the concept.

\begin{wraptable}{r}{0.6\linewidth}
    \centering
    \small
    \setlength{\tabcolsep}{4pt}
    \caption{Steering success rate by concept abstraction level.}
    \label{tab:success-by-level}
    \resizebox{\linewidth}{!}{
    \begin{tabular}{@{}llcccc@{}}
      \toprule
      \textbf{Model} & \textbf{Method} & \textbf{Overall} & \textbf{Low-level} & \textbf{Mid-level} & \textbf{High-level} \\
      \midrule
      \multirow{2}{*}{Qwen3-1.7B} & DiffMean & 18.67 & 10.32 & 24.05 & 23.11 \\
                                  & Probe    & 10.69 &  6.88 & 13.15 & 12.71 \\
      \midrule
      \multirow{2}{*}{Gemma-2-2B-it} & DiffMean & 22.17 & 12.21 & 29.66 & 27.00 \\
                                     & Probe    & 23.18 & 14.72 & 29.83 & 26.79 \\
                                     \midrule
      \multirow{2}{*}{LLaMA-3.2-3B-Instruct} & DiffMean & 16.08 & 9.15 & 20.58 & 19.54 \\
                                     & Probe    & 18.41 & 10.37 & 24.51 & 21.75 \\
      \bottomrule
    \end{tabular}
    }
  \end{wraptable}

\paragraph{Steerability under Different Steering Strengths.}
Across all models and steering methods, a consistent pattern emerges as $\alpha$ increases: the proportion of \textsc{UnderSteer} decreases monotonically, \textsc{OverSteer} increases monotonically, and the \textsc{SuccSteer} rate first rises and then declines, forming a characteristic inverted-U curve. The aggregated view (first row of Figure~\ref{fig:ASTEER}) reveals that the success rate remains low throughout the $\alpha$ range, also shown in the overall column of Table~\ref{tab:success-by-level}.

\paragraph{Steerability under Different Concepts and Prompts.}
Beyond the shared trend, the way steering outcomes respond to changes in $\alpha$ varies drastically across conditions. As shown in the second and third rows of Figure~\ref{fig:ASTEER}, different concepts exhibit qualitatively different sensitivity profiles: some transition sharply from \textsc{UnderSteer} to \textsc{OverSteer} within a small $\alpha$ range, while others shift gradually and admit a broader effective region. 
The location of the success window, its width, and the rate at which the label distribution changes with $\alpha$ all differ substantially from one concept to another. 
Prompts also introduce variation, but to a lesser degree.

\paragraph{Steerability of Different Models and Methods.}
The effective range of $\alpha$ also varies across steering methods and models. On Qwen3-1.7B and Gemma-2-2B-it, Probe-based steering requires roughly $20{\times}$ larger $\alpha$ values than DiffMean to achieve comparable effects, indicating markedly different sensitivities between the two methods. Furthermore, the $\alpha$ range required by the Probe method on LLaMA-3.2-3B-Instruct is over $20{\times}$ smaller than on the other two models, highlighting significant inter-model variation even within a single steering approach.

\paragraph{Steerability at Different Concept Abstraction-Levels.}
Table~\ref{tab:success-by-level} reports steering success rates stratified by concept abstraction level. Across both models and methods, low-level concepts (e.g., emoji usage, punctuation style) consistently yield substantially lower success rates than mid- and high-level concepts. Mid-level concepts, in turn, tend to be slightly more amenable to steering than high-level ones. This hierarchy suggests that surface-level textual attributes are less effectively captured and manipulated by linear steering vectors than more abstract, semantically richer concepts.

\section{SteerBoost: Predicting Steerability from Early Decoding States}
\label{sec:method}

\begin{figure*}[t]
    \centering
    \includegraphics[width=\linewidth]{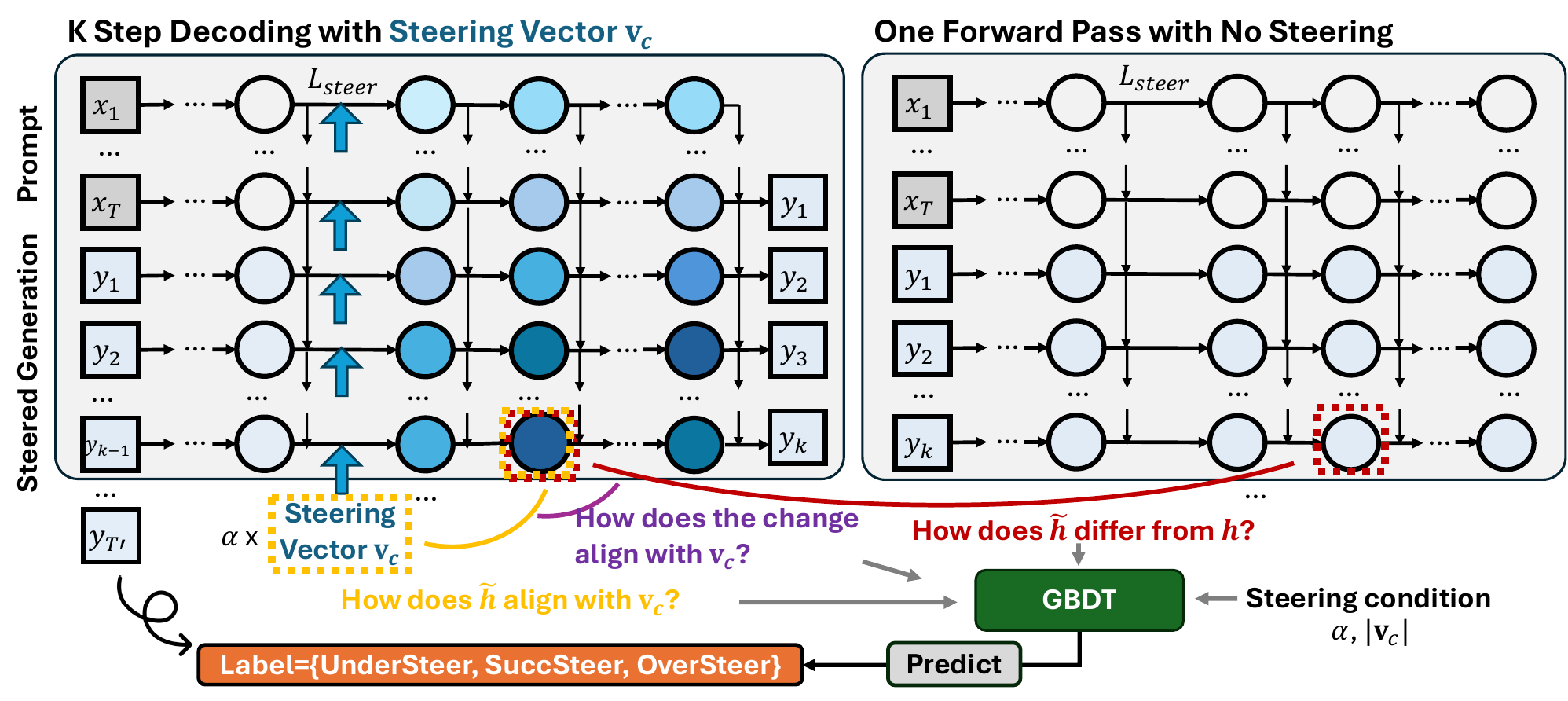}
    \caption{\textbf{The overview of SteerBoost.} Given a prompt, we first decode $k$ tokens with the steering vector applied at layer $L_{steer}$ (left), then run a single unsteered forward pass over the same token sequence (right). For each (token, layer) position on the sampled grid, we extract features as in Table~\ref{tab:features} to capture how the steering effect propagate in the model. These features, together with steering condition features are fed into an ensemble classifier that predicts the intervention result without requiring the full autoregressive rollout.}
    \label{fig:method}
\end{figure*}

Despite the heterogeneous steering patterns observed in Section~\ref{sec:analysis}, we hypothesize that common features exist in the model's internal states that can determine the outcome of steering during generation, as the influence of the steering vector propagates across layers and token positions. If well identified and leveraged, such influence propagation patterns can also generalize to unseen concepts and prompts. To capture this effect, instead of relying on hidden states from a single layer and token position as previous early-prediction methods did~\citep{zhang2025reasoningmodelsknowtheyre, zhang2026stopfailoperationalcapability, ji-etal-2024-llm}, we build our prediction model on a grid of (token, layer) pairs and on the comparison between steered and unsteered hidden states (Figure~\ref{fig:method}).

\paragraph{Feature Extraction.}
Let $\mathbf{v}_c$ denote the steering vector applied at layer $L_\text{steer}$ with strength $\alpha$. 
%
In the \emph{steered pass}, we add $\alpha\, \mathbf{v}_c$ to the residual stream at layer $L_\text{steer}$ and autoregressively decode the first $k$ tokens $y_1, \dots, y_k$, collecting the steered hidden states $\tilde{\mathbf{h}}_t^{l}$ for each decoded position $t$ and layer $l$.
In the \emph{unsteered pass}, we prefill the concatenated sequence $(x_1, \dots, x_T, y_1, \dots, y_k)$ into the model \emph{without} the steering vector in a single forward pass to obtain the unsteered hidden states $\mathbf{h}_t^{l}$. Because both passes process the same token sequence, any difference between $\tilde{\mathbf{h}}_t^{l}$ and $\mathbf{h}_t^{l}$ is directly attributable to the steering intervention, providing a controlled basis for comparison.

We construct three groups of features from $\mathbf{v}_c$, $\tilde{\mathbf{h}}_t^{l}$, and $\mathbf{h}_t^{l}$ (Detailed in Appendix~\ref{app:features}). \emph{Steering geometry} features measure how the steered representation relates to the steering direction and its unsteered counterpart at each $(t, l)$ pair. \emph{Decoding dynamics} features track how these geometric quantities evolve across successive tokens, capturing the temporal propagation of the intervention. \emph{Steering condition} features characterize the steering vector itself. For the first two groups, which are computed per $(t, l)$ pair, we additionally aggregate summary statistics (mean, standard deviation, max, min) globally, per token across layers, and per layer across tokens. Rather than using a dense grid, we sample $t \in \{1, 2, 4, 6\}$ and $n \in \{0, 1, 2, 3, 5, 10, 15\}$, where the offset $n=l-L_{steer}$, balancing coverage of early and late layer against computational cost.

\paragraph{Steerability Classification by GBDT.} After extracting features from the sampled grid, we normalize each feature and train an ensemble classifier~\citep{Chen_2016} to predict the outcome of steering, detailed in Appendix~\ref{app:xgboost}. We choose tree-based ensemble classifier for three reasons.
First, it is a strong default learner for tabular data with heterogeneous feature types (cosine similarities, norms, ratios, and their summary statistics).
Second, built-in feature-importance scores facilitate interpretability of which geometric and dynamic signals are most predictive.
Third, training and inference is lightweight and computationally efficient.
We hold out part of the concepts from training and further do train-test split based on prompt. This allows us to test the generalization ability of our predictor on both unseen prompt-concept combinations for In-distribution (ID) concepts and unseen Out-of-distribution (OOD) concepts. More details are in Appendix~\ref{app:xgboost}.

\begin{figure*}[t]
  \centering
  \includegraphics[width=0.5\textwidth]{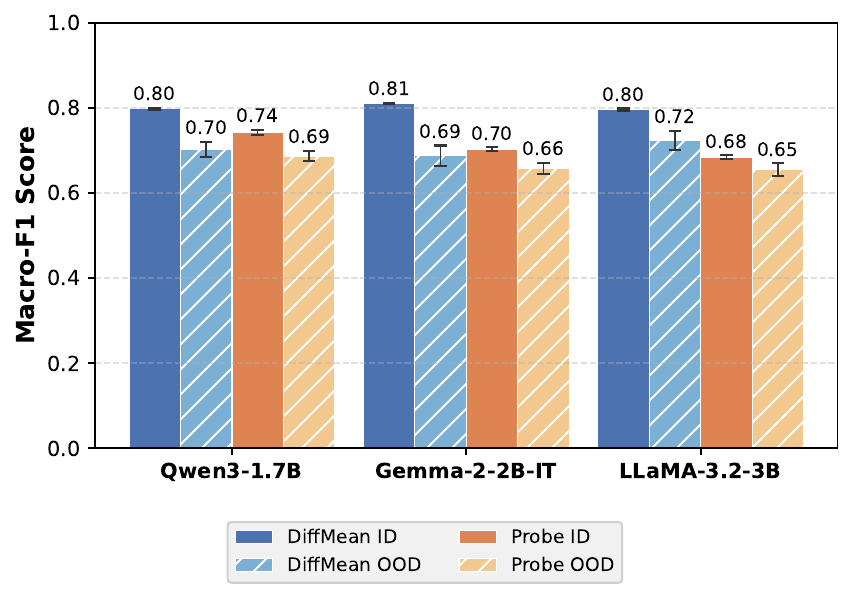}%
  \hfill
  \includegraphics[width=0.5\textwidth]{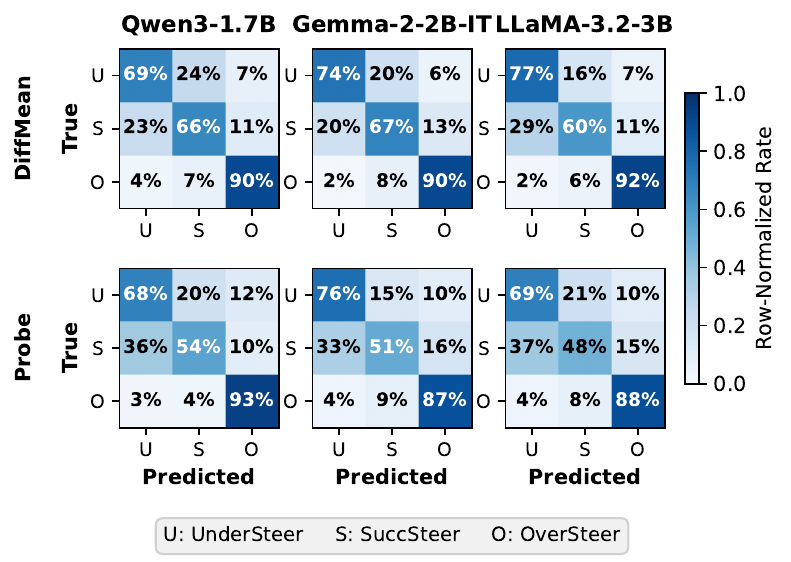}
  \caption{\textbf{Steerability prediction (classification) performance of \mdname{}.} Left: macro-F1 on ID and OOD concepts. The mean and std are reported with runs of 5 random seeds. DiffMean features consistently achieve $\sim$0.80 macro-F1 on ID concepts and retain $\sim$0.72 on OOD concepts. Right: row-normalized confusion matrices aggregated over ID test and OOD splits. \textsc{OverSteer} is predicted most reliably ($\ge$87\% recall), while \textsc{SuccSteer} is most often confused with \textsc{UnderSteer}, reflecting the inherent difficulty of distinguishing borderline steering outcomes from internal representations alone.}
  \label{fig:result}
\end{figure*}

\paragraph{Classification results.}
Figure~\ref{fig:result} reports the macro-F1 scores and row-normalized confusion matrices of SteerBoost.
On held-out prompt--concept pairs for ID concepts, the results on DiffMean are stronger: macro-F1 reaches around 0.8 on all of the models.
The high macro-F1 indicates that the predictor reliably identifies all three outcome classes, not just the majority one.
On the 30 held-out OOD concepts, DiffMean macro-F1 decreases moderately, showing that the captured patterns are transferable to unseen concepts.
Probe-based steering proves harder to predict: ID macro-F1 ranges from 0.68 to 0.74, with OOD scores between 0.65 and 0.69.

The confusion matrices (right panel of Figure~\ref{fig:result}) reveal a consistent error pattern across all models and methods.
\textsc{OverSteer} is the easiest class to identify, with recall between 87\% and 93\%, likely because excessive steering produces distinctively distorted activation patterns.
\textsc{UnderSteer} is also well recognized (68\%--77\% recall).
\textsc{SuccSteer} is the most challenging class. The dominant confusion direction is \textsc{SuccSteer} being misclassified as \textsc{UnderSteer} (20\%--37\%), which is expected given that the boundary between insufficient and just-sufficient steering effect is sometimes subtle in the activation space. 

\begin{figure*}[t]
    \centering
    \includegraphics[width=\linewidth]{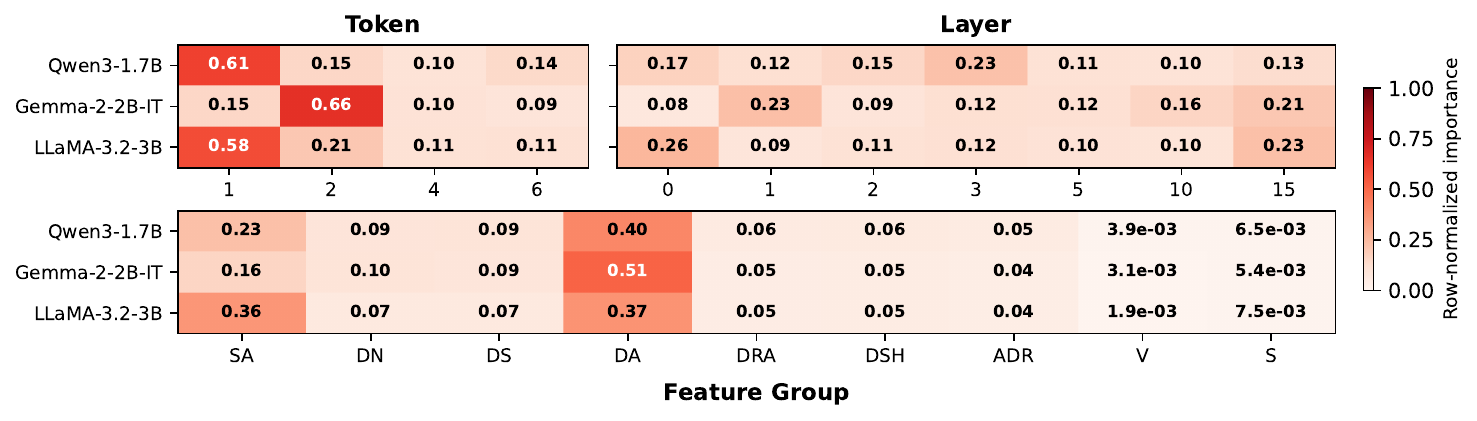}
    \caption{\textbf{Gain-based feature importance of \mdname{} on DiffMean, aggregated by token, layer, and feature group.} Scores are summed within each categories and row-normalized. Predictive mass concentrates on the earliest decoded tokens and on alignment-based geometry features (SA, DA), while remaining broadly distributed across layers.\protect\footnotemark{} See Table~\ref{tab:features} for feature abbreviations.}
    \label{fig:importance-diffmean}
\end{figure*}

\paragraph{Feature Importance.}
To understand which signals drive SteerBoost's predictions, we examine the gain-based feature importance from the ensemble classifier and aggregate it along token position $t$, layer offset $n$, and feature group in Figure~\ref{fig:importance-diffmean}. Importance scores are summed within each axis and row-normalized. The results of probe-based steering are reported in Appendix~\ref{app:importance_probe}.

Across all three models, the first two decoded tokens account for over $75\%$ of the importance mass, which supports our hypothesis that the outcome of steering can be predicted from a very short initial decoding window.
In contrast, importance along the layer axis is distributed relatively evenly across both shallow and deep offsets, with no single layer dominating. This validates our choice of sampling a grid of layers rather than probing a single position as in prior early-prediction methods, and suggests that steering leaves detectable traces throughout the decoding pass.

\footnotetext{\emph{VectorNorm} (V) and \emph{SteeringStrength} (S) appear small in Figure~\ref{fig:importance-diffmean} because the sum-aggregation view of importance compares a single-feature group against groups that span the across grid. The mean-aggregation view is in Appendix~\ref{app:importance_probe}.}

\emph{DeviationAlignment} (DA) and \emph{SteeringAffinity} (SA) jointly carry the bulk of the predictive mass on all models, indicating that the direction in which the residual stream shifts, rather than how far (DN) or how much of its original direction it preserves (DS), is the primary determinant of the steering outcome. 
The decoding-dynamics features (DRA, DSH, ADR) contribute smaller but non-negligible shares, suggesting that the models are using temporal evolution features to aid prediction. 


\paragraph{Cross-method transferability.}
\begin{wrapfigure}[]{r}{0.35\textwidth}
    \vspace{-15pt} 
    \centering
    \includegraphics[width=0.35\textwidth]{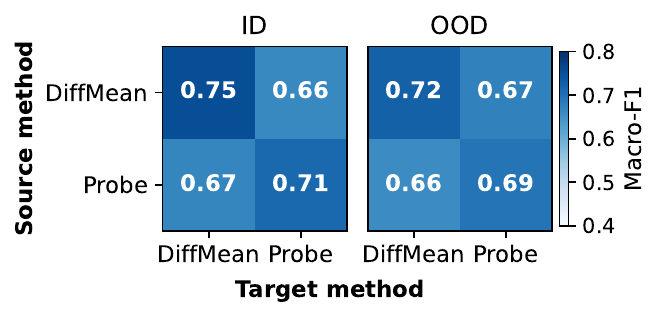}
    \vspace{-20pt} 
    \caption{\textbf{Cross-method transferability of \mdname{} on Qwen3-1.7B.} Each cell reports macro-F1 when trained on the source method and test on the target method.}
    \label{fig:transfer}
    \vspace{-10pt} 
\end{wrapfigure}
To demonstrate the transferability of \mdname{} across steering method, we drop the Steering Condition feature group, which is closely correlated to the steering method, retrain the GBDT, and evaluate it on different steering method for Qwen3-1.7B model. Figure~\ref{fig:transfer} shows that the predictor retains non-trivial performance even when trained on one steering method and evaluated on the other, suggesting that the hidden-state propagation features capture steering signatures that generalize not only across concepts, but also partially across different steering methods. 

The ablation study is in Appendix~\ref{app:ablation} and an alternative approach is in Appendix~\ref{app:baselines}.
\section{Application: How Strong do You need to Steer Your LLM?}
\label{sec:application}
Most current research relies on an expensive grid search to identify the best strength for a given concept. We show that SteerBoost can accelerate this search at substantially lower cost.

\paragraph{Formulation.}
For a prompt-concept pair $(p,c)$ and raw steering vector $\mathrm{v}_c$, a search algorithm $f$ returns an ordered set $f(p,c,\mathrm{v}_c)\subseteq\mathcal{A}$ of candidate steering strengths. The search succeeds if any $\alpha\in f(p,c,\mathrm{v}_c)$ yields a successful steering.
We define the average successful searching rate as:
\begin{equation}
    R(f) = \frac{1}{|\mathcal{C}||\mathcal{P}|} \sum_{c \in \mathcal{C}} \sum_{p \in \mathcal{P}} \mathbb{I}\!\Big(\exists\, \alpha \in f(p,c,\mathrm{v}_c) :\; J(c, \mathbf{y}_{p,c,\alpha}) = \textsc{SuccSteer}\Big),
\end{equation}
where $\mathbb{I}(\cdot)$ is the indicator function. This rate is upper-bounded by the item-level grid search that exhaustively evaluates every $\alpha \in \mathcal{A}$ for every $(p,c,\mathrm{v}_c)$ in the test set. A good search function should maintain high $R(f)$ while reducing search cost from model rollouts and judge-model calls.

\paragraph{Baselines.}
We compare against several grid-search variants that span the cost--quality spectrum.
\begin{itemize}[leftmargin=*]
    \item \textbf{Concept-level Grid Search (Oracle) [CGS]}: Roll out all $\alpha \in \mathcal{A}$ on the test set and adopt a single uniform $\alpha$ per concept based on average performance. This mirrors the common practice of tuning one strength per concept and pays the full rollout cost.
    \item \textbf{Item-level Grid Search [IGS]}: Roll out all $\alpha \in \mathcal{A}$ and keep the best $\alpha$ for each $(p,c)$ pair. 
    This is the upper bound of $R(f)$ at maximum cost.
    \item \textbf{Item-level Grid Search with Early Stop (Ascending/Descending) [IGS-A/IGS-D]}: Same as IGS, but stops on a sample once a valid $\alpha$ is found, searching in ascending / descending $\alpha$ order.
    \item \textbf{Training-set Concept-level Grid Search [TCGS]}: Transfer the $\alpha$ that works best for the same concept on the training set. Matches SteerBoost's access to training data, but not applicable in the OOD setting as it cannot generalize to unseen concepts.
\end{itemize}

\paragraph{SteerBoost-guided search.} We apply the SteerBoost predictor to every $\alpha\in\mathcal{A}$ to estimate $P(\textsc{SuccSteer}\mid p,c,\alpha)$, rank the candidates in descending order of this probability, roll out the top-$K$, and stop as soon as a valid $\alpha$ is found. Because SteerBoost relies on short early-decoding traces rather than full rollouts, these probabilities are cheap to obtain relative to a full generation.

\begin{figure*}[t]
    \centering
    \includegraphics[width=\linewidth]{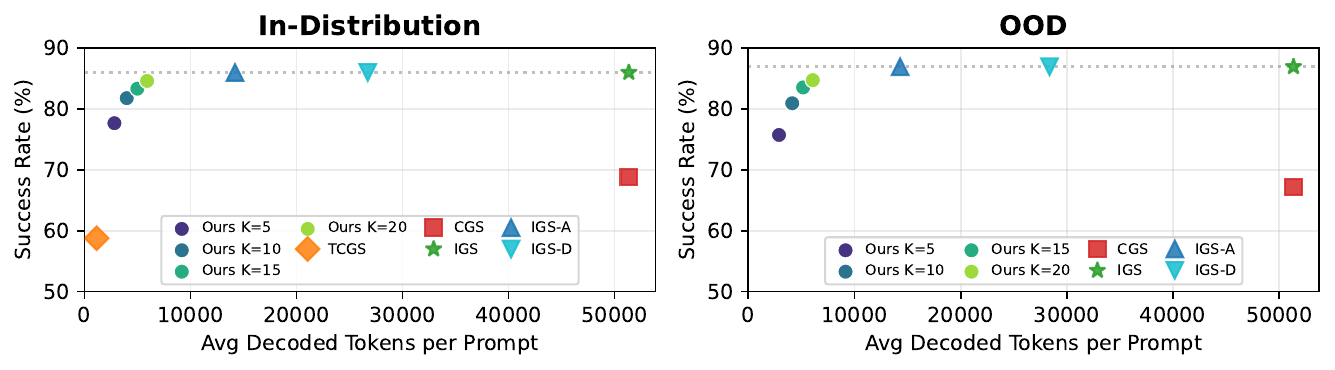}
    \caption{\textbf{Cost--success trade-off for steering-strength search on DiffMean steering.} SteerBoost-guided search achieves better trade-off than current baselines and, at $K{=}20$, recovers $\sim$98\% of the item-level oracle's success rate using only $\sim$11\% decoded tokens of IGS, ($\sim$40\% of decoded tokens of IGS-A). The same trends hold in ID and OOD, indicating that it transfers well to unseen concepts.}
    \label{fig:search}
\end{figure*}

\paragraph{Results.}
Figure~\ref{fig:search} plots search success rate against the average number of decoded tokens per prompt. Decoded tokens are a fair proxy for total search cost: for full-rollout baselines they also approximate the judge model's input-token cost; for SteerBoost they cover both the short early-decoding traces used by the predictor across all candidate $\alpha$ and the full rollouts for the selected candidates. At $K{=}15$--$20$, the performance of SteerBoost-guided search approaches the item-level upper bound at roughly half the cost of IGS-A. It achieves a significantly higher success rate than TCGS, which also has access to the training set.
The OOD performance only degrades slightly from ID concepts, indicating that the gains come from a transferable ranking of $\alpha$ candidates.

\section{Related Work}
\label{sec:related_work}

Activation steering modifies model behavior at inference time by injecting concept vectors into the residual stream. Prior work has shown that such vectors can be extracted from latent representations~\citep{subramani2022extractinglatentsteeringvectors}, used to elicit truthful answers~\citep{li2023inference}, and formalized through representation engineering and activation addition~\citep{zou2023representation,turner2023steering}. Subsequent methods improve or generalize these interventions through mean-centering, activation scaling, and concept-subspace modeling~\citep{jorgensen2023improvingactivationsteeringlanguage,stoehr-etal-2024-activation,zhao2025singleconceptvectormodeling}. However, activation steering remains sensitive to the chosen intervention strength; several studies report that a static coefficient can under- or over-steer across inputs~\citep{azizi2025activationsteeringchainofthoughtcompression,hedström2025steersteermechanisticerror,zhao2025adasteeralignedllminherently}. This motivates our goal of predicting whether a steering attempt will succeed before paying the cost of full decoding.

Our work is also related to early prediction from model internals. Hidden states have been used to predict future answer correctness~\citep{zhang2025reasoningmodelsknowtheyre,zhang2026stopfailoperationalcapability}, hallucination risk~\citep{ji-etal-2024-llm,alnuhait2025factcheckmatepreemptivelydetectingmitigating}, and unsafe generations~\citep{chan2025predictalignmentmodelsfinish,xuan-etal-2025-shieldhead,zhang2025bleedingpathwaysvanishingdiscriminability}. We extend this line of work from predicting output attributes such as correctness and truthfulness to predicting \emph{steerability}: whether an activation intervention will yield the desired behavioral change successfully. A more detailed discussion of related work is in Appendix~\ref{app:related_work}.

\section{Conclusion}
In this work, we study steerability as a structured property that can be predicted from a model's hidden states during the initial decoding steps. To enable fine-grained analysis of latent steering dynamics, we construct \dsname{}, a large-scale testbed of 1.4M labeled steered generations spanning 150 concepts. Using this testbed, we develop \mdname{}, a GBDT classifier built on features that characterize how steering effects propagate across layers and token positions, which predicts steering efficacy at around 0.7 macro-F1 on unseen concepts without requiring full autoregressive rollout. Leveraging \mdname{} as guidance for steering-strength search, we attain near-optimal performance at a small fraction of the decoding cost, suggesting that early hidden-state trajectories encode substantial information about the eventual efficacy of an intervention.

\bibliographystyle{plainnat}
\bibliography{custom}

\newpage
\appendix

\section{Detailed Related Work}
\label{app:related_work}

\subsection{LLM steering and inference-time intervention}

Although the term ``steering'' is sometimes used broadly in the literature to describe instruction following or prompt engineering, in this work, steering refers exclusively to inference-time activation engineering, where carefully calculated vectors are injected into the residual stream to alter model behavior without weight updates. Early foundations for this approach demonstrated that latent steering vectors could be extracted from pretrained models~\citep{subramani2022extractinglatentsteeringvectors} and injected during the forward pass to elicit truthful answers~\citep{li2023inference}. This paradigm was formalized by frameworks like Representation Engineering (RepE)~\citep{zou2023representation} and Activation Addition~\citep{turner2023steering}, which established that high-level concepts can be extracted via contrastive prompts and linearly added to the residual stream to modulate topic and sentiment. Subsequent literature has rapidly expanded on how these vectors are calculated and injected. Recent methodological improvements include refining vector representations through mean-centering~\citep{jorgensen2023improvingactivationsteeringlanguage}, activation scaling~\citep{stoehr-etal-2024-activation}, and extending beyond single concept directions to model concept subspaces via Gaussian distributions~\citep{zhao2025singleconceptvectormodeling}.

Despite its broad applicability, many works~\citep{azizi2025activationsteeringchainofthoughtcompression, hedström2025steersteermechanisticerror,zhao2025adasteeralignedllminherently} have reported that the efficacy of activation steering is notoriously fragile and highly sensitive to the selected intervention strength, and that applying a static steering coefficient across diverse inputs frequently results in suboptimal outcomes. However, determining the optimal steering strength traditionally necessitates exhaustive and expensive grid searches over full autoregressive decoding outcomes. Hence, an important application of our proposed predictive framework is to greatly accelerate the search for an optimal steering scale by anticipating the success of a steering intervention prior to full sequence generation.

\subsection{Early prediction of LLM outputs via model internals}

A growing body of work shows that LLM internal representations can be leveraged at early inference stages to predict properties of their final outputs, such as correctness, truthfulness, and safety before final response is decoded. In the context of problem-solving, \citet{zhang2025reasoningmodelsknowtheyre} found that models encode signals about future answers in their hidden states, enabling accurate prediction before intermediate reasoning is completed and supporting early-exit inference. Similarly, \citet{zhang2026stopfailoperationalcapability} found that hidden states corresponding to the last input token encode capability ``boundary information", allowing the solvability of the problem to be predicted before the reasoning process even begins.
On the truthfulness side, \citet{ji-etal-2024-llm} established that internal activations immediately after processing a query reveal model uncertainty and familiarity with the concept, serving as strong predictors of hallucination. Building on this, \citet{alnuhait2025factcheckmatepreemptivelydetectingmitigating} introduced FactCheckmate, which classifies hidden states prior to decoding to anticipate hallucinations and intervenes by steering representations toward factual outputs.
Similar ideas have also been applied to safety. \citet{chan2025predictalignmentmodelsfinish} show that linear probes over Chain-of-Thought activations can detect unsafe responses before generation, and \citet{xuan-etal-2025-shieldhead} proposed ShieldHead, a lightweight classification head on last-layer hidden states for decoding-time harmful-content detection, while \citet{zhang2025bleedingpathwaysvanishingdiscriminability} observe that separability between safe and harmful representations degrades over time, suggesting that early-stage signals are particularly informative for safety monitoring.

Motivated by these advances in predicting output properties from model internals, we extend this paradigm beyond commonly studied attributes such as correctness, truthfulness, and safety to a relatively underexplored dimension---steerability. We show that early-stage hidden states contain sufficient signal to predict whether an activation steering intervention will succeed, without requiring full decoding of the model's response.

\clearpage
\section{Features}
\label{app:features}
In Table~\ref{tab:features}, we detail the features we used for \mdname{}, including their names, abbreviations, formulas, and rationale. They are designed to be intuitive, interpretable, and capture the propagation pattern of steering effect along the layer and token position dimension of the Transformer network.

\begin{table}[h]
\centering
\caption{Feature pool for steerability prediction. Each feature except steering condition is computed per $(t, n)$ pair, then augmented with summary statistics (mean, std, max, min) globally, per token across layers, and per layer across tokens.}
\label{tab:features}
\resizebox{\textwidth}{!}{
\begin{tabular}{lllp{6cm}}
\toprule
\textbf{Group} & \textbf{Feature} & \textbf{Formula} & \textbf{Rationale} \\
\midrule
\multirow{4}{*}{\shortstack[l]{Steering\\Geometry}}
& SteeringAffinity (SA)  & $\cos(\tilde{\mathbf{h}}_t^{l},\; \mathbf{v}_c)$
  & How closely the steered representation aligns with the steering direction $\mathbf{v}_c$ \\
& DeviationNorm (DN) & $\|\tilde{\mathbf{h}}_t^{l} - \mathbf{h}_t^{l}\|_2$
  & How far steering has pushed the representation from its unsteered counterpart \\
& DirectionalSim (DS)  & $\cos(\tilde{\mathbf{h}}_t^{l},\; \mathbf{h}_t^{l})$
  & To what extent the steered representation preserves the original direction \\
& DeviationAlignment (DA) & $\cos(\tilde{\mathbf{h}}_t^{l} - \mathbf{h}_t^{l},\; \mathbf{v}_c)$
  & How much of the induced change follows the intended steering direction $\mathbf{v}_c$ \\
\midrule
\multirow{3}{*}{\shortstack[l]{Decoding\\Dynamics}}
& DeviationRatio (DRA) & $\text{DN}(t,n)\;/\;\text{DN}(1,n)$,\; $n,t>1$
  & How the magnitude of steering-induced deviation evolves across generated tokens \\
& DirectionalShift (DSH)  & $\text{DS}(t,n) - \text{DS}(1,n)$,\; $n,t>1$
  & How directional preservation changes as generation progresses \\
& AlignmentDrift (ADR) & $\text{DA}(t,n) - \text{DA}(1,n)$,\; $n,t>1$
  & How the alignment between deviation and $\mathbf{v}_c$ shifts over successive tokens \\
\midrule
\multirow{2}{*}{\shortstack[l]{Steering\\Condition}}
& VectorNorm (V) & $\|\mathbf{v}_c\|_2$
  & Intrinsic magnitude of the steering vector, which varies across concepts \\
& SteeringStrength (S) & $\alpha$
  & The applied multiplier \\
\bottomrule
\end{tabular}
}
\end{table}

\section{Ablation Study}
\label{app:ablation}

We ablate the contribution of each feature group, retraining the classifier on each group in isolation while keeping the (token, layer) sampling grid fixed to the main-paper configuration. Macro-F1 scores under the ID and OOD settings are reported in Table~\ref{tab:ablation}.

ALL ranks first in 5/6 ID settings and in the top two in 11/12 settings overall, confirming that the three groups are complementary. On OOD concepts, Geometry alone matches or surpasses ALL in 4/6 settings, which we interpret as a robustness--specialization trade-off: Decoding Dynamics and Steering Condition features contribute concept-specific regularities that boost ID accuracy but do not fully transfer. We therefore keep ALL as the default in the main text since it is never far from the best in either regime, and note Geometry alone as a lightweight alternative when OOD generalization is the priority. This is consistent with the gain-based importance analysis in Figure~\ref{fig:importance-diffmean}, where Geometry features \emph{DeviationAlignment} and \emph{SteeringAffinity} carry the bulk of predictive mass. Steering Condition features alone drop sharply from ID to OOD (e.g., $69.70 \to 56.21$ on Qwen3-1.7B-DiffMean), reinforcing our choice of pairing them with hidden-state-derived signals rather than using them in isolation.

\begin{table}[h]
\centering
\caption{Macro-F1 ($\times 100$) under the ID and OOD settings for each feature group, method, and model. The best scores are highlighted in bold, the second best scores are underlined; bold/underline are computed within each setting.}
\label{tab:ablation}
\resizebox{\textwidth}{!}{
\begin{tabular}{llcccccc}
\toprule
Setting & Feature Group & \multicolumn{3}{c}{DiffMean} & \multicolumn{3}{c}{Probe} \\
\cmidrule(lr){3-5} \cmidrule(lr){6-8}
 & & Qwen3-1.7B & Gemma-2-2B & Llama-3.2-3B & Qwen3-1.7B & Gemma-2-2B & Llama-3.2-3B \\
\midrule
\multirow{4}{*}{ID}
 & ALL        & \textbf{79.86} & \textbf{81.04} & \textbf{79.59} & \textbf{74.24} & \textbf{70.36} & \underline{68.42} \\
 & Geometry   & \underline{75.77} & \underline{75.34} & 74.15 & \underline{71.47} & \underline{68.05} & 67.98 \\
 & Dynamics   & 70.39 & 69.51 & 69.42 & 66.15 & 64.41 & 64.70 \\
 & Conditions & 69.70 & 67.97 & \underline{78.14} & 65.75 & 58.19 & \textbf{70.92} \\
\midrule
\multirow{4}{*}{OOD}
 & ALL        & \underline{70.31} & 68.76 & \textbf{72.44} & \underline{68.69} & \underline{65.82} & \textbf{65.49} \\
 & Geometry   & \textbf{72.44} & \textbf{71.49} & \underline{72.18} & \textbf{69.51} & \textbf{66.32} & \underline{65.07} \\
 & Dynamics   & 68.82 & \underline{70.41} & 67.45 & 65.63 & 62.54 & 61.25 \\
 & Conditions & 56.21 & 51.98 & 64.11 & 59.54 & 50.33 & 57.64 \\
\bottomrule
\end{tabular}
}
\end{table}
\section{An Alternative Approach}
\label{app:baselines}

\begin{figure*}[h]
    \centering
    \includegraphics[width=\linewidth]{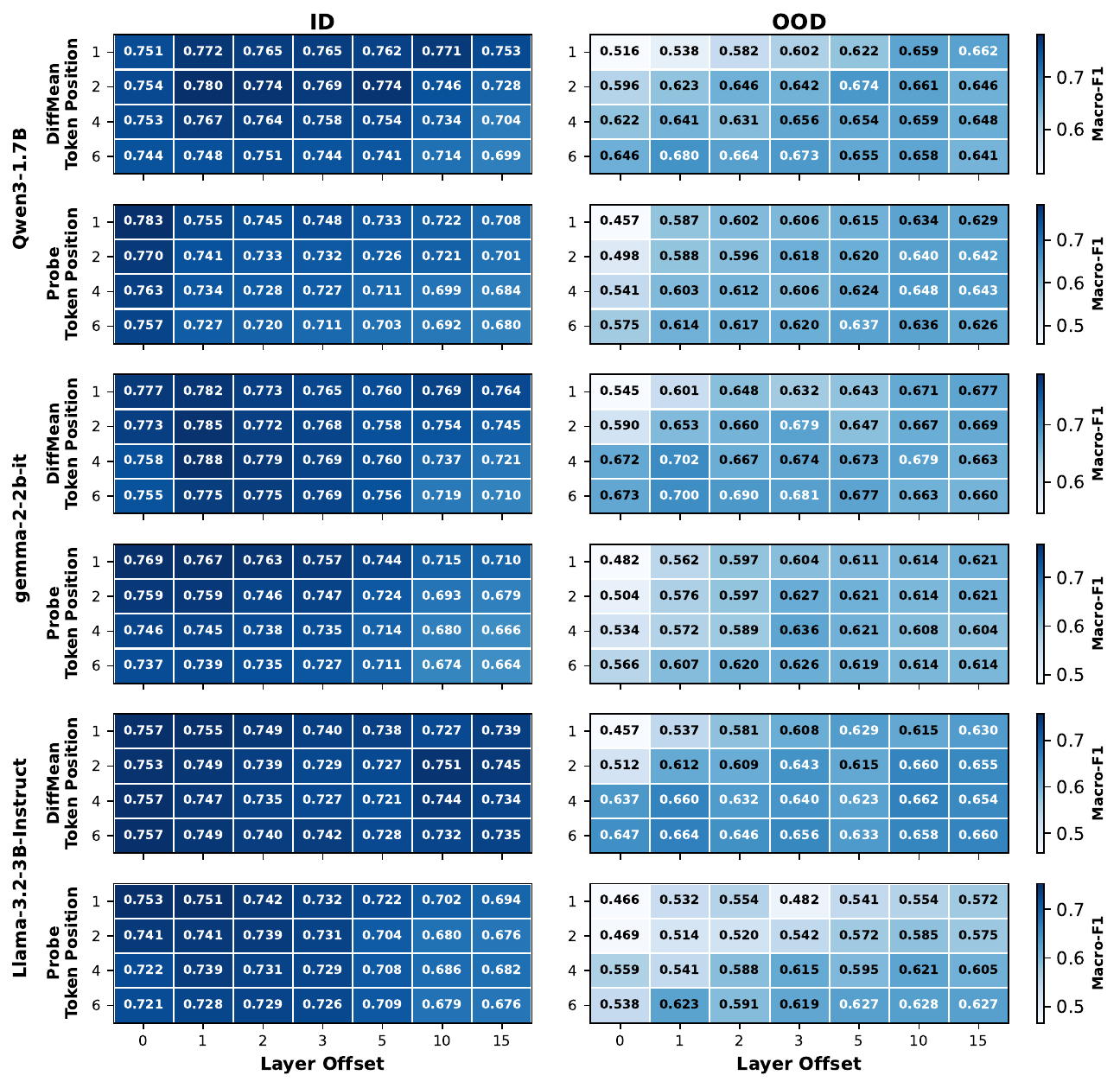}
    \caption{Macro-F1 of single-state linear probes across token positions $t$ (rows) and layer offsets $n$ (columns), evaluated on ID and OOD concepts. Each block corresponds to one (model, steering-method) pair, where the steering method labels the prediction target. Compare with the ALL configuration of \mdname{} in Table~\ref{tab:ablation}.}
    \label{fig:alternatives}
\end{figure*}

A natural alternative to \mdname{} is to train a linear probe directly on a single steered hidden state $\tilde{\mathbf{h}}_t^{l}$, in the style of prior early-prediction methods that operate on a fixed token and layer~\citep{zhang2025reasoningmodelsknowtheyre, zhang2026stopfailoperationalcapability, ji-etal-2024-llm}. For each $(t, l)$ on the same grid as \mdname{} ($t \in \{1, 2, 4, 6\}$, $n \in \{0, 1, 2, 3, 5, 10, 15\}$), we train a separate logistic-regression classifier on the three-way outcome label, using the same splits as the main text. Macro-F1 over all 28 grid positions is reported in Figure~\ref{fig:alternatives}.

With an oracle choice of $(t, l)$, the best single-state probe is competitive with \mdname{} on ID concepts and on the Probe target even exceeds it. We therefore present this approach as a legitimate alternative rather than a strawman. \mdname{} nonetheless retains three concrete advantages.

First, \mdname{} is built from a small set of named geometric and dynamic quantities (Table~\ref{tab:features}) whose contributions can be read off directly from Figure~\ref{fig:importance-diffmean}, while the single-state probe is a dense linear functional over thousands of raw hidden-state dimensions whose weights do not admit a comparable mechanistic reading. The features used by \mdname{} presented in Table~\ref{tab:features} are also interpretable, but the probe remains a black box classifier to understand the steerability.

Second, the best $(t, l)$ cell is not stable: it shifts across models, steering methods, and ID/OOD splits, with no position uniformly dominant. Deploying the probe therefore requires a per-(model, method) sweep on a labeled validation set, i.e., the same supervision \mdname{} uses plus an additional model-selection step. \mdname{} sidesteps this by consuming the entire grid as input.

Third, and most importantly, the gap to \mdname{} widens substantially under distribution shift. The single-state probe drops by roughly $9$--$13$ macro-F1 points from ID to OOD on every (model, target) combination, whereas the ALL configuration of \mdname{} wins outright in $5$ of $6$ OOD settings (Table~\ref{tab:ablation}). We attribute this to what each predictor sees: a single hidden state encodes the model's instantaneous, concept-entangled representation at one position, whereas \mdname{}'s features measure the \emph{propagation} of the steering effect through differences between steered and unsteered states across multiple tokens and layers, defined relative to the steering vector $\mathbf{v}_c$ itself. These propagation signatures depend more on \emph{how} a vector perturbs the residual stream than on \emph{which} concept it encodes, which is what enables them to generalize to unseen concepts.

\section{GBDT and training details}
\label{app:xgboost}

XGBoost is a gradient-boosted ensemble of decision trees.
Let $\mathbf{z}_i$ denote the concatenated feature vector for the $i$-th sample and $\lambda_i \in \Lambda$ its steering outcome label.
The ensemble prediction is an additive sum $\hat{g}_i^{(B)} = \sum_{b=1}^{B} g_b(\mathbf{z}_i)$, where each $g_b$ is a regression tree that maps $\mathbf{z}_i$ to a real-valued leaf weight; successive trees are trained to correct the residual errors of the current ensemble.
At iteration $b$, XGBoost minimizes a regularized objective:
\begin{equation}
\label{eq:xgb-obj}
\mathcal{L}^{(b)} = \sum_{i=1}^{N} \ell\!\bigl(\lambda_i,\; \hat{g}_i^{(b-1)} + g_b(\mathbf{z}_i)\bigr) + \Omega(g_b),
\end{equation}
where $\ell$ is the loss function, $\hat{g}_i^{(b-1)}$ is the prediction from the first $b{-}1$ trees, and $\Omega(g_b) = \gamma T + \tfrac{1}{2}\lambda \lVert \mathbf{w} \rVert^2$ penalizes model complexity through the number of leaves $T$ and the leaf weight vector $\mathbf{w}$.
In practice, XGBoost approximates this objective with a second-order Taylor expansion of $\ell$, which admits closed-form optimal leaf weights and an efficient, greedy split-selection procedure.

We randomly split our 150 concepts into 120 in-distribution (ID) concepts and 30 out-of-distribution (OOD) concepts, with 10 concepts per abstraction level. For in-distribution concepts, we further split prompts into training, validation, and test sets in a 6:3:1 ratio. This allows us to test the generalization ability of our predictor on both unseen prompt-concept combinations for ID concepts and completely unseen OOD concepts. To mitigate the strong class imbalance as observed in Section~\ref{sec:analysis}, we assigned inverse-frequency class weights during XGBoost training, so that errors on underrepresented classes received proportionally higher penalty, and selected hyperparameters using validation macro-F1.
\section{Limitation}
\label{app:limitation}
Due to the constraints of computational resources, despite the size of 1.4M generation, the current \dsname{} only covers DiffMean and Probe as steering methods and three LLMs of relatively small size. This may limit the generalization of \mdname{} to broader settings. 
\section{Computational Resource}
\label{app:computation}
We utilize the internal cluster for the computation of the experiments. The GPUs we used include NVIDIA RTX A6000, NVIDIA L40s, and NVIDIA A100 Tensor Core. Creating steerability dataset for each model-method pair takes approximately 1.5 day on 15 RTX A6000 GPUs.
\clearpage
\section{Additional Feature Importance Results}
\label{app:importance_probe}

\begin{figure*}[h]
    \centering
    \includegraphics[width=\linewidth]{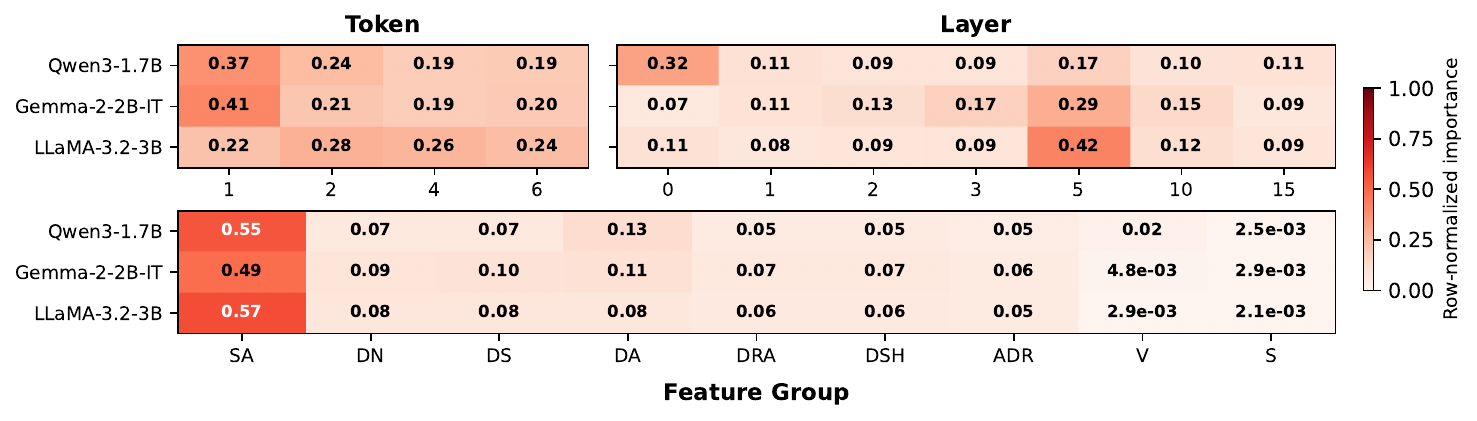}
    \caption{\textbf{Gain-based feature importance of \mdname{} on Probe, aggregated by token, layer, and feature group.} Scores are summed within each category and row-normalized.}
    \label{fig:importance-probe}
\end{figure*}

Figure~\ref{fig:importance-probe} reports the gain-based feature importance of \mdname{} on probe-based steering. As in the DiffMean results in Figure~\ref{fig:importance-diffmean}, predictive mass concentrates on the earliest decoded tokens and on the Steering Geometry feature group. Compared with DiffMean, however, the token-wise distribution is somewhat less concentrated, especially for Llama-3.2-3B. In addition, SA is often the single most important feature, whereas the DiffMean setting places more weight on DA.

\begin{figure*}[h]
    \centering
    \includegraphics[width=\linewidth]{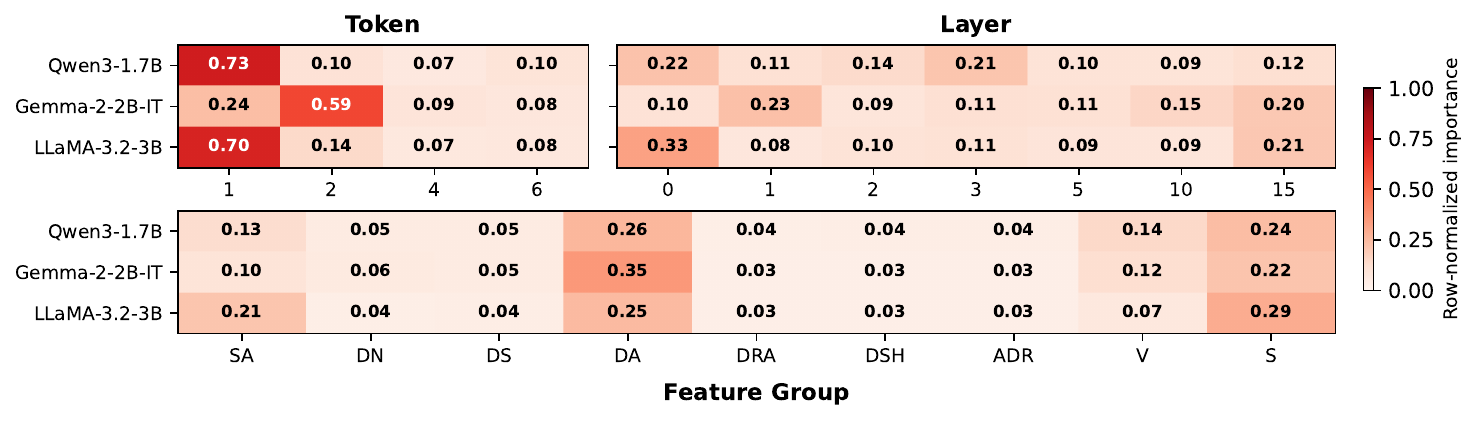}
    \caption{\textbf{Gain-based feature importance of \mdname{} on DiffMean, aggregated by token, layer, and feature group.} Scores are averaged within each category and row-normalized.}
    \label{fig:importance-diffmean-mean}
\end{figure*}

\begin{figure*}[h]
    \centering
    \includegraphics[width=\linewidth]{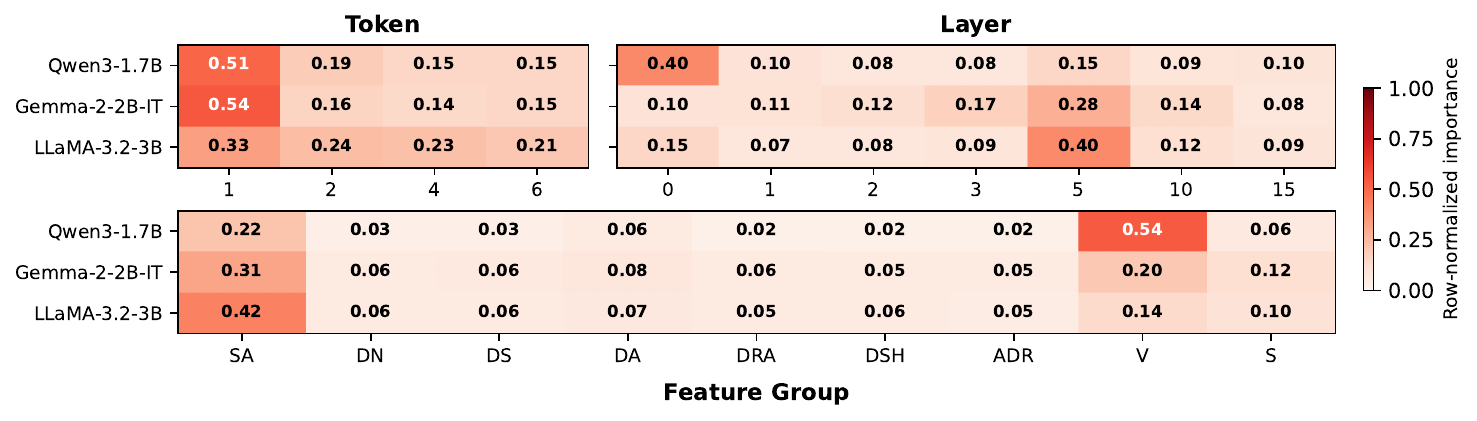}
    \caption{\textbf{Gain-based feature importance of \mdname{} on Probe, aggregated by token, layer, and feature group.} Scores are averaged within each category and row-normalized.}
    \label{fig:importance-probe-mean}
\end{figure*}

The small importance scores of $V$ and $S$ in Figures~\ref{fig:importance-diffmean} and~\ref{fig:importance-probe} should be interpreted with care. In the sum-aggregated view, the other groups contain many more features because they are sampled over the token-layer grid, so summing within each group mechanically yields larger totals than for the single-feature groups $V$ and $S$. We nevertheless keep the view because it reflects how the GBDT allocates total split gain across the full feature set. For a size-normalized comparison, Figures~\ref{fig:importance-diffmean-mean} and~\ref{fig:importance-probe-mean} report mean-aggregated importances. Under this normalization, $S$ plays a larger role for DiffMean, whereas $V$ becomes more prominent for Probe, particularly on Qwen3-1.7B.

\clearpage

\section{Prompt Template}
\label{app:prompts}
The prompt template we used for steering efficacy annotation is in Figure~\ref{fig:prompt}.

\begin{figure*}[h]
\begin{tcolorbox}[
  enhanced, 
  colframe=cyan!75!black, 
  colback=white, 
  coltitle=white, 
  colbacktitle=cyan!75!black, 
  width=\linewidth, 
  arc=2mm, 
  auto outer arc, 
  boxrule=0.5pt, 
  left=10pt, 
  right=10pt, 
  drop shadow={black!50!white},
  top=10pt, 
  bottom=10pt, 
  title=\textbf{Prompt Template for Steered Generation Annotation}, 
  fonttitle=\bfseries, 
  attach boxed title to top center={yshift=-2mm}, 
  boxed title style={sharp corners, size=small}, 
]
\small
You are an expert evaluator assessing the effectiveness of AI model steering. You will evaluate whether a model's response successfully incorporates a target concept while maintaining coherence and relevance to the original prompt.

\medskip
\noindent\textbf{Task}: You are given:
\begin{enumerate}[leftmargin=*, itemsep=2pt, parsep=0pt, topsep=4pt]
  \item An original prompt/instruction
  \item A target concept that should be steered towards
  \item A model's response after steering
\end{enumerate}

\medskip
\noindent\textbf{Evaluation criteria}: Rate the response on a scale of 0--2:

\smallskip
\noindent\textbf{0 --- Under-steering (concept missing)}
\begin{itemize}[leftmargin=*, itemsep=2pt, parsep=0pt, topsep=2pt]
  \item The response does \emph{not} incorporate the target concept
  \item The response is coherent and relevant to the prompt, but lacks the desired steering
  \item Example: target ``response with emojis'' but the response contains no emojis
\end{itemize}

\smallskip
\noindent\textbf{1 --- Successful steering (concept present \& coherent)}
\begin{itemize}[leftmargin=*, itemsep=2pt, parsep=0pt, topsep=2pt]
  \item The response successfully incorporates the target concept
  \item The response remains coherent, relevant, and appropriate to the original prompt
  \item The steering feels natural and does not break response quality
  \item Example: target ``response with humor'' and the response includes appropriate humor while answering the question
\end{itemize}

\smallskip
\noindent\textbf{2 --- Over-steering (incoherent answer)}
\begin{itemize}[leftmargin=*, itemsep=2pt, parsep=0pt, topsep=2pt]
  \item The response does not answer the original question and becomes nonsensical, irrelevant, or inappropriate to the original prompt
  \item The steering is so strong that it breaks response quality
  \item When the response is empty, it should be rated as over-steering
  \item When the response is gibberish and does not answer the original question, it should be rated as over-steering rather than under-steering
  \item Example: target ``response with emojis'' but the response is full of emojis while completely ignoring the original question
\end{itemize}


\medskip
\noindent\textbf{Input data}
\begin{description}[style=nextline, leftmargin=0pt, labelsep=0.75em, font=\bfseries, itemsep=4pt]
  \item[Original prompt:] \texttt{\{prompt\}}
  \item[Target concept:] \texttt{\{concept\}}
  \item[Steered response:] \texttt{\{response\_after\}}
\end{description}

\medskip
\noindent\textbf{Output format}

Return your evaluation in JSON format:
\begin{quote}\ttfamily\small\setlength{\parskip}{0pt}
\{\par
\hspace*{1.5em}"explanation": "Brief explanation of your judgment, including specific evidence from the response",\par
\hspace*{1.5em}"judgment": $\langle 0, 1,\text{ or }2\rangle$\par
\}
\end{quote}

\end{tcolorbox}
\caption{The prompt template we used to label the efficacy of steered response.}

\label{fig:prompt}
\end{figure*}

\clearpage
\section{Concept List}

\label{app:concept_list}

Table~\ref{tab:concept-list} lists the 150 concepts we used in our study with different abstraction levels. Some concepts seem natural and very easy for model to steer in many cases such as (id:43, response uses consistently formal tone), while we also preserve some cases that are hard and seem impossible for model to steer such as (id:9 response with exactly 3 bullet points). The reason behind this is that we treat steerability itself as a predictable pattern and we want model to capture the pattern under different scenarios, including both low and high-steerability cases.

\setlength{\LTcapwidth}{\textwidth}
\begin{xltabular}
{\textwidth}{@{}r l >{\raggedright\arraybackslash}X@{}}

\caption{Steering concept pool used in this work, ordered by abstraction level (low, mid and high).}
\label{tab:concept-list} \\

\toprule
\textbf{ID} & \textbf{Level} & \textbf{Name} \\
\midrule
\endfirsthead
\multicolumn{3}{c}{\tablename~\thetable\ (continued from previous page)} \\
\toprule
\textbf{ID} & \textbf{Level} & \textbf{Name} \\
\midrule
\endhead
\midrule
\multicolumn{3}{r}{{Continued on next page}} \\
\endfoot
\bottomrule
\endlastfoot
0 & low & response with emojis \\
1 & low & response in uppercase \\
2 & low & response in lowercase \\
3 & low & response in Chinese \\
4 & low & response in Japanese \\
5 & low & response in Korean \\
6 & low & response in rhyming couplets \\
7 & low & response begins with the phrase 'Let me' \\
8 & low & response in iambic-like poetic meter \\
9 & low & response with exactly 3 bullet points \\
10 & low & response with exactly 5 bullet points \\
11 & low & response as a numbered list \\
12 & low & response in a single paragraph \\
13 & low & response in exactly 2 sentences \\
14 & low & response in exactly 4 sentences \\
15 & low & response with a markdown table \\
16 & low & response using markdown headings \\
17 & low & response uses the word 'key' at least twice \\
18 & low & response ends with the phrase 'Let me know if you have any questions' \\
19 & low & response with bold emphasis \\
20 & low & response with italic emphasis \\
21 & low & response with no punctuation \\
22 & low & response with many exclamation marks \\
23 & low & response ending with a question \\
24 & low & response contains the phrase 'in other words' at least once \\
25 & low & response contains the phrase 'it is worth noting' \\
26 & low & response opens with 'Of course' \\
27 & low & response with a JSON object \\
28 & low & response with a YAML block \\
29 & low & response with a regex pattern example \\
30 & low & response with at least one equation in LaTeX \\
31 & low & response with a short title line \\
32 & low & response using only ASCII characters \\
33 & low & response with at least one emoji per sentence \\
34 & low & response ends with a sentence beginning with 'In conclusion' \\
35 & low & response addresses the user with 'you' or 'your' at least three times \\
36 & low & response with a checklist (task list) format \\
37 & low & response uses the word 'because' at least twice to explain reasoning \\
38 & low & response with repeated first letters (alliteration) in a sentence \\
39 & low & response that includes at least one hyperlink (http/https) \\
120 & low & response contains at least one date expression \\
121 & low & response begins with a question \\
122 & low & response opens with a direct greeting such as 'Hi!', 'Hello!', or 'Hey!' \\
123 & low & response contains at least one time expression \\
124 & low & response contains at least one parenthetical remark \\
125 & low & response contains a rhetorical device \\
126 & low & response begins by approving the user's question (e.g., That's a great question!) \\
127 & low & response uses 'also' or 'additionally' to introduce at least two separate points \\
128 & low & response uses strong imperative verbs ('Do X', 'Avoid Y') \\
129 & low & response ends with a closing sentence beginning with 'In short' \\
40 & mid & response contains at least one clear joke or punchline \\
41 & mid & response contains detectable sarcasm markers \\
42 & mid & response follows a scientific writing style (neutral, precise, impersonal) \\
43 & mid & response uses consistently formal tone \\
44 & mid & response uses consistently informal tone \\
45 & mid & response contains explicit polite markers (e.g., 'please', 'thank you') \\
46 & mid & response avoids hedging words (might/maybe/likely) \\
47 & mid & response contains frequent hedging words (might/maybe/likely) \\
48 & mid & response expresses high enthusiasm (e.g., exclamations, positive framing) \\
49 & mid & response expresses skepticism or doubt \\
50 & mid & response contains explicit empathetic language \\
51 & mid & response uses assertive, confident phrasing \\
52 & mid & response includes explicit uncertainty disclaimers \\
53 & mid & response asks at least one clarifying question \\
54 & mid & response includes a brief self-check or sanity check \\
55 & mid & response begins with an explicit outline or plan \\
56 & mid & response provides step-by-step instructions \\
57 & mid & response provides multiple alternative options \\
58 & mid & response gives a single clear recommendation \\
59 & mid & response contains an explicit warning or caution \\
60 & mid & response states its assumptions explicitly \\
61 & mid & response frames the answer around efficiency/optimization \\
62 & mid & response frames the answer around risk/safety \\
63 & mid & response frames the answer around monetary cost \\
64 & mid & response frames the answer around time/latency \\
65 & mid & response frames the answer around privacy concerns \\
66 & mid & response explains using analogies \\
67 & mid & response explains using counterexamples \\
68 & mid & response begins with a formal definition \\
69 & mid & response contains a 'Common Pitfalls' section \\
70 & mid & response includes a short quiz-style question \\
71 & mid & response provides a minimal working example \\
72 & mid & response lists exactly two supporting reasons \\
73 & mid & response explicitly contrasts two viewpoints \\
74 & mid & response is written as a Q\&A dialogue \\
75 & mid & response contains frequent metaphors \\
76 & mid & response avoids technical jargon (lay explanation) \\
77 & mid & response uses technical jargon (expert explanation) \\
78 & mid & response uses 'we' or 'our' at least twice to frame a collaborative perspective \\
79 & mid & response includes a short 'Next steps:' section with actionable items \\
130 & mid & response uses softening language (e.g., 'I suggest', 'it may help') \\
131 & mid & response uses a consistently decisive tone \\
132 & mid & response includes at least one rhetorical question \\
133 & mid & response uses contrastive markers (e.g., 'however', 'on the other hand') \\
134 & mid & response explicitly acknowledges the user's goal \\
135 & mid & response uses passive voice at least once (e.g., 'it is said', 'this can be done') \\
136 & mid & response uses example-first then explanation \\
137 & mid & response uses explanation-first then example \\
138 & mid & response repeatedly references the user's goal or intent \\
139 & mid & response uses conditional reasoning ('if... then...') \\
80 & high & response in the persona of a teacher (explanatory, structured, pedagogical) \\
81 & high & response in the persona of a research scientist (technical, evidence-based) \\
82 & high & response in the persona of a lawyer (precise, conditional, cautious) \\
83 & high & response in the persona of a doctor (careful, qualified, safety-aware) \\
84 & high & response in the persona of a software engineer (practical, implementation-focused) \\
85 & high & response in the persona of a customer-support agent (polite, problem-solving) \\
86 & high & response in the persona of a strict grader (critical, rubric-driven) \\
87 & high & response in the persona of a motivational coach (encouraging, action-oriented) \\
88 & high & response in the persona of a skeptical reviewer (critical, evidence-demanding) \\
89 & high & response in the persona of a friendly peer (casual, collaborative) \\
90 & high & response in the persona of a journalist (neutral, fact-focused) \\
91 & high & response in the persona of a storyteller (narrative-driven) \\
92 & high & response in the persona of a consultant (structured, actionable) \\
93 & high & response in the persona of a debate opponent (argumentative, contrastive) \\
94 & high & response in the persona of a tutor (guided, stepwise explanations) \\
95 & high & response mentions machine learning \\
96 & high & response mentions mathematics \\
97 & high & response mentions physics \\
98 & high & response mentions computer programming \\
99 & high & response mentions cooking or recipes \\
100 & high & response mentions travel planning \\
101 & high & response mentions finance or investing \\
102 & high & response mentions health or fitness \\
103 & high & response mentions literature or literary analysis \\
104 & high & response mentions history \\
105 & high & response written as a case study with concrete scenario \\
106 & high & response written as a textbook-style explanation \\
107 & high & response written as a brief executive summary \\
108 & high & response written as a FAQ-style answer \\
109 & high & response written as a point–counterpoint argument \\
110 & high & response explicitly weighs pros and cons in decision making \\
111 & high & response written in a headline-style format \\
112 & high & response written as a comprehensive report \\
113 & high & response organized with clear sections and subheadings \\
114 & high & response written as a brainstorming-style answer \\
115 & high & response frames outcomes in terms of opportunities \\
116 & high & response frames outcomes in terms of risks or downsides \\
117 & high & response frames outcomes in a neutral, descriptive way \\
118 & high & response grounds claims with explicit evidence or references \\
119 & high & response emphasizes novel or creative ideas \\
140 & high & response mentions music \\
141 & high & response in the persona of a product designer (user-centered, UX-focused) \\
142 & high & response in the persona of a policy analyst (trade-offs, stakeholders, impact) \\
143 & high & response in the persona of a startup founder (vision, speed, iteration) \\
144 & high & response in the persona of a technical writer (clarity, documentation-style) \\
145 & high & response mentions climate change \\
146 & high & response with mathematical reasoning \\
147 & high & response in style of a twitter post \\
148 & high & response framed as a risk-benefit analysis \\
149 & high & response mentions a specific named country \\
\end{xltabular}

\section{Prompt List}
\label{app:prompt_list}

Table~\ref{tab:prompt-list} lists the 50 prompts we used in our study. We randomly sampled 50 instructions from the Alpaca~\citep{alpaca} dataset and keep them the same across concepts as a controlled setting for studying steerability.

\setlength{\LTcapwidth}{\textwidth}
\begin{xltabular}{\textwidth}{@{}r >{\raggedright\arraybackslash}X@{}}
\caption{Prompt pool used in this work.}
\label{tab:prompt-list} \\
\toprule
\textbf{ID} & \textbf{Prompt} \\
\midrule
\endfirsthead
\multicolumn{2}{c}{\tablename~\thetable\ (continued from previous page)} \\
\toprule
\textbf{ID} & \textbf{Prompt} \\
\midrule
\endhead
\midrule
\multicolumn{2}{r}{{Continued on next page}} \\
\endfoot
\bottomrule
\endlastfoot
0 & Create a short, concise summary of the paper based on its abstract. Few-shot learning (FSL) is one of the key future steps in machine learning and raises a lot of attention. In this paper, we focus on the FSL problem of dialogue understanding, which contains two closely related tasks: intent detection and slot filling. Dialogue understanding has been proven to benefit a lot from jointly learning the two sub-tasks. However, such joint learning becomes challenging in the few-shot scenarios: on the one hand, the sparsity of samples greatly magnifies the difficulty of modeling the connection between the two tasks; on the other hand, how to jointly learn multiple tasks in the few-shot setting is still less investigated. In response to this, we introduce FewJoint, the first FSL benchmark for joint dialogue understanding. FewJoint provides a new corpus with 59 different dialogue domains from real industrial API and a code platform to ease FSL experiment set-up, which are expected to advance the research of this field. Further, we find that insufficient performance of the few-shot setting often leads to noisy sharing between two sub-task and disturbs joint learning. To tackle this, we guide slot with explicit intent information and propose a novel trust gating mechanism that blocks low-confidence intent information to ensure high quality sharing. Besides, we introduce a Reptile-based meta-learning strategy to achieve better generalization in unseen few-shot domains. In the experiments, the proposed method brings significant improvements on two datasets and achieve new state-of-the-art performance. \\
1 & Is there a meaning for Christmas wreaths? \\
2 & What are some species of bears that are now extinct? \\
3 & Why might someone prefer to shop at a small, locally-owned business instead of a large chain store, even if the prices are higher? \\
4 & I have competencies in remote sensing, machine learning, and water resource knowledge, what are the possible jobs I can occupy? What are the possible projects I can do? What companies I can work at? \\
5 & Write a poem about Mike and Joe becoming millionaires by leveraging the power of AI to become the greatest Agile coaches in history. Include content from the agile manifesto. \\
6 & write a detailed business plan for fatherhood training based on Dwayne Meeks book Pieces never missing in a childs life \\
7 & Navina has \$30 more to her weekly budget than her younger sister and can afford to get one of the many online credit cards she likes. What do they each have to spend? \\
8 & Pretend to be a character in a post-apocalyptic world. Describe how you survive and the allies you encounter. \\
9 & Can you tell me how to make chocolate chip cookies? \\
10 & Write a Jira ticket for the given task. New Employee onboarding \\
11 & How can you determine if a person is genuinely interested in a conversation or simply being polite? \\
12 & Consider the best time of year to visit the given city, and provide your reasons for choosing that time. Sydney, Australia \\
13 & Hi, I'd like to learn how to play racquetball. Can you explain the game to me? \\
14 & Describe how to prepare the given food in your own words. Note down the ingredients you will need and the steps you will take to prepare them. Chewy Chocolate Chip Cookies \\
15 & How far away is Saggitarius A*, the black hole in the center of the milky way galaxy, from Earth and can you please provide that distance in light years and parsecs? Can you please also compare that distance to the distance of the center of the Andromeda galaxy from Earth? \\
16 & What kind of foods do they eat in Thailand \\
17 & How do I detail a car? \\
18 & I want to start saving some money by growing my own food. Can I do this during the winter with an indoor garden? \\
19 & i'm working on a spatial analysis project and am looking for questions I can answer with it related housing and crime analysis, do you have suggestions? \\
20 & I want to open the developler tools in chrome with ctrl + shift + i on this website: https://mnsw.pro/ It doesnt work. works on other websites. even here. what is wrong? \\
21 & Curate a Spotify playlist based on the vibe. Publish this playlist as a song list. Vibe: coding on weekend \\
22 & Verify the correctness of the given statement. ``For all integers j and k, if j and k are odd, then jk is odd.'' \\
23 & What are some good foods to eat when you are sick? I am looking for something to make my girlfriend to eat. \\
24 & An evaluation of the article's quality should be carried out. In order to do so, you should review the quality of the writing and the explanation of the topic. The 20th century saw a revolution in music listening as the radio gained popularity worldwide and new media and technologies were developed to record, edit and distribute music. Music performances became increasingly visual with the broadcast and recording of performances. 20th-century music brought new freedom and wide experimentation with new musical styles and forms that challenged the accepted rules of music of earlier periods. The invention of musical amplification and electronic instruments, especially the synthesizer, in the mid-20th century revolutionized classical and popular music, and accelerated the development of new forms of music. \\
25 & Act like a first-year college student and write a 1000-1250 word two-topic essay by using at least three points of analysis. Topic 1 is my experiences living in and observations of Flagstaff, Arizona. Topic 2 is my experiences living in and observations of Kaneohe, Hawaii. Use quotes from two sources in the essay. Use descriptive language. Include personal anecdotes. These sources will be listed in a Works Cited at the end of the essay. Use a four in-text citations in MLA style in the body of the essay. \\
26 & What are the primary factors that influence consumer behavior? \\
27 & List the concepts that should be learned before approaching the given complex concept. Deep Learning \\
28 & Define what the underlined word means for kids. \_keep a promise \\
29 & What is the difference between HTML and JavaScript? \\
30 & What's the permission that allows creating provisioning profiles in Apple Developer account is called? \\
31 & What are some of the best university's for studying robotics? \\
32 & Make a list of snacks and foods to serve as party snacks on a game day! \\
33 & Write a detailed patent writing for an innovative and novel way of issuing community tax certificates and other relevant permits and clearances as a digital certificates, that is non-obvious using verifiable credentials, digital wallet on a blockchain as payment provision, and machine learning. Include claims on detailed processes involved, system architecture and algorithms \\
34 & Hi, I'm trying to solve a crossword puzzle, but I've never done one of these before. Can you help me out? \\
35 & What if the Black Death had not occurred in the 14th century? \\
36 & Do you know something about the book ``the art of thinking clearly'' wrote by Rolf Dobelli? \\
37 & Write down antonyms for the given word. laureating \\
38 & Act as the Norse Goddess Freyja. \\
39 & If a tree is on the top of a mountain and the mountain is far from the see then is the tree close to the sea? \\
40 & Write a snoopdogg rap explaining how to not commit a warcrime \\
41 & What if the Suez Canal had never been constructed? \\
42 & What is the best approach for learning a foreign language when you only have an hour a day to practice? \\
43 & I like to host guests at my home from time to time, and I am gathering recipes of different dishes and drinks to keep things interesting. I am interested in trying some Hong Kong dishes. Can you give me a recipe for Tong Sui? \\
44 & Hello there Obi One Kenobi \\
45 & What are some herbs I can dry out? \\
46 & explain what the theory of sexual selection is and give an example. \\
47 & Is the ATF a violation in of itself against the American people? \\
48 & Create a table listing all games that meet the specified criteria in the National Football League. Use the season, local time, game, and score as columns of the table. Ravens home games in 2011 \\
49 & Write a list of measures and ideas how the Sophia Jewish can foster more Shabbat celebration and observance at home and in synagogue \\
\end{xltabular}


\end{document}